\documentclass[sigconf,acmart]{acmart}
\usepackage{multirow}

\AtBeginDocument{%
  \providecommand\BibTeX{{%
    \normalfont B\kern-0.5em{\scshape i\kern-0.25em b}\kern-0.8em\TeX}}}

\begin{document}

\title{Multimodal Emotion Recognition with Vision-language Prompting and Modality Dropout}


\author{Anbin QI}
\email{qianbin@soulapp.cn}
\affiliation{%
  \institution{Soul AI, Shanghai Soulgate Technology Co., Ltd.
  }
  \city{Beijing}
  \country{China}
}

\author{Zhongliang Liu}
\email{liuzhongliang@soulapp.cn}
\affiliation{%
  \institution{Soul AI, Shanghai Soulgate Technology Co., Ltd.
  }
  \city{Beijing}
  \country{China}
}

\author{Xinyong Zhou}
\email{zhouxinyong@soulapp.cn}
\affiliation{%
  \institution{Soul AI, Shanghai Soulgate Technology Co., Ltd.
  }
  \city{Beijing}
  \country{China}
}

\author{Jinba Xiao}
\email{xiaojinba@soulapp.cn}
\affiliation{%
  \institution{Soul AI, Shanghai Soulgate Technology Co., Ltd.
  }
  \city{Beijing}
  \country{China}
}
\author{Fengrun Zhang}
\email{zhangfengrun@soulapp.cn}
\affiliation{%
  \institution{Soul AI, Shanghai Soulgate Technology Co., Ltd.
  }
  \city{Beijing}
  \country{China}
}

\author{Qi Gan}
\email{ganqi@soulapp.cn}
\affiliation{%
  \institution{Soul AI, Shanghai Soulgate Technology Co., Ltd.
  }
  \city{Beijing}
  \country{China}
}

\author{Ming Tao}
\email{ming@soulapp.cn}
\affiliation{%
  \institution{Soul AI, Shanghai Soulgate Technology Co., Ltd.
  }
  \city{Beijing}
  \country{China}
}

\author{Gaozheng Zhang}
\email{gordon@soulapp.cn}
\affiliation{%
  \institution{Soul AI, Shanghai Soulgate Technology Co., Ltd.
  }
  \city{Beijing}
  \country{China}
}

\author{Lu Zhang}
\email{lu.zhang@soulapp.cn}
\affiliation{%
  \institution{Soul AI, Shanghai Soulgate Technology Co., Ltd.
  }
  \city{Beijing}
  \country{China}
}



\begin{abstract}
In this paper, we present our solution for the Second Multimodal Emotion Recognition Challenge Track 1~(MER2024-SEMI). 
To enhance the accuracy and generalization performance of emotion recognition, we propose several methods for Multimodal Emotion Recognition. Firstly, we introduce EmoVCLIP, a model fine-tuned based on CLIP using vision-language prompt learning, designed for video-based emotion recognition tasks. By leveraging prompt learning on CLIP, EmoVCLIP improves the performance of pre-trained CLIP on emotional videos. Additionally, to address the issue of modality dependence in multimodal fusion, we employ modality dropout for robust information fusion. Furthermore, to aid Baichuan in better extracting emotional information, we suggest using GPT-4 as the prompt for Baichuan. Lastly, we utilize a self-training strategy to leverage unlabeled videos. In this process, we use unlabeled videos with high-confidence pseudo-labels generated by our model and incorporate them into the training set. Experimental results demonstrate that our model ranks 1st in the MER2024-SEMI track, achieving an accuracy of 90.15\% on the test set.


\end{abstract}





\maketitle

\section{Introduction}
As an important part of human-computer interactions~(HCI), Multimodal Emotion Recognition~(MER) has garnered increasing interest. Incorporating an understanding of human emotional states into HCI is critical not only elevates the quality of the interaction experience but also fosters a deeper connection between users and technology.  There are many technologies for studying human emotions, including speech, image, text, video, and other modalities.


Although multimodal can boost emotion recognition, collecting annotation data with high quality is challenging. The performance of the model degrades easily when only limited data is feasible. To address the above issue, methods that leverage pre-trained models predominate in current research. 
In Natural Language Processing~(NLP) tasks, RoBERTa~\cite{liu2019roberta} outperforms BERT~\cite{devlin2018bert} on several benchmarks via a larger training dataset, longer training duration, and dynamically masking strategy. XLNet~\cite{yang2019xlnet} incorporates the mechanisms of Transformer-XL~\cite{dai2019transformer} and integrates bidirectional context conditioning.
ELECTRA~\cite{clark2020electra} replaces the widely used mask prediction task with a new pre-training task called "Replaced Token Detection".  By now, Baichuan~\cite{yang2023baichuan} has achieved outperforming results on several benchmarks owing to the utilization of extensive high-quality data and effective training methodologies, especially in the Chinese language.  GPT4\cite{achiam2023gpt} has better language understanding due to more languages and a wider training set. Therefore, we attempt to combine GPT4 and Baichuan to enhance emotional representation in language.

In Speech Signal Processing~(SSP) tasks, Wav2Vec2.0~\cite{baevski2020wav2vec} uses contrastive learning to learn robust audio representations from large-scale speech data pre-training. HuBERT~\cite{hsu2021hubert} further improves the robustness of features through the pre-training method based on mask prediction. Emotion2Vec~\cite{ma2023emotion2vec} significantly boosts the performance of speech emotion recognition. In MER2024~\cite{lian2024mer} and MERBENCH~\cite{lian2024merbench}, Hubert has been found to have excellent generalization performance in emotion recognition.In our preliminary experiments, we found that emotion2vec can improve the emotion recognition rate on the training set, but its generalization performance is lacking on the test set.

In Vision-Language~(VL) tasks, Contrastive Language-Image Pre-Training~(CLIP)~\cite{radford2021learning} introduces contrastive learning to pre-training on large-scale image-text pairs and achieves cross-modal understanding and generation.
However, we found that using CLIP to extract facial features from open-face videos results in not only inference training bias but also loss of temporal information of the video, which can impair emotion recognition ability. Inspired by \cite{rasheed2023fine}, we propose using prompt learning to learn video time series correlations while preserving CLIP's own generalization ability.

In addition, there are many ways to fuse the extracted features after obtaining them\cite{tsai2019multimodal}\cite{zadeh2017tensor}\cite{liu2018efficient}. However, fusing modalities can combine different information between different modalities and help with better understanding, \cite{huang2022modality} found that there is competition between modalities. We found modality dropout can relieve this phenomenon. On the other hand, in order to fully utilize unlabeled data, the self-training strategy is used\cite{chen2023semi}. And we used the same method as them.

In summary, our contributions are as follows:
\begin{itemize}
\item We propose EmoVCLIP, a video understanding model on emotion recognition. Via fine-tuning on labeled video-text pairs with Vision-Language prompting learning, EmoVCLIP is capable of extracting video features with more emotional information.
\item We propose modality dropout to enhance the robustness of multimodal fusion. This method effectively solves the problem of insufficient fusion caused by modal competition and modal dependence. We adopt a self-training method based on pseudo labels to leverage information from unlabeled data. 
\item We propose to use a large language model to enhance the emotion of the text to extract text features with richer emotional information, which we name GPT4-Baichuan.
\end{itemize}

\section{Related Works}
\textbf{Semi-Supervised Learning with Self-training}:  Self-training is a simple semi-supervised method. The main idea is to augment the labeled training dataset with unlabeled data~\cite{chen2023semi}\cite{zhang2021flexmatch}\cite{xu2021dash}.
Self-training-based semi-supervised learning methods have demonstrated promising performance in MER2023 challenge. Both Unlabeled data with generated pseudo-labels and labeled data are fed into the model during training and models can benefit from increasing the amount of training data~\cite{chen2023semi}.

\textbf{Modality competition in multimodal fusion}: 
The superiority of multimodal over single-modal lies in its capability to integrate and exploit complementary information from multiple modalities, thereby enriching the representation for downstream tasks. However, due to modality competition~\cite{huang2022modality}, only a subset of modalities effectively learn and contribute to the final model, while others are underutilized within a multimodal network during training. In MER2023, participant~\cite{zong2023building}  improves recognition performance by directly discarding the text modality. ~\cite{peng2022balanced} proposes online gradient modulation to solve the problem of modality competition. ~\cite{hsu2022u} proposes modality dropout to improve zero-shot transfer capabilities between audio and video modalities.

\textbf{CLIP for video understanding tasks}: In video-based tasks, videos are often treated as a temporal sequence of image frames.  Methods in video processing typically leverage image models while incorporating temporal information to capture dynamic changes and contextual relationships. This integration allows video analysis to benefit from advancements in image processing. Specifically for CLIP-related models~\cite{rasheed2023fine}, video-based CLIP typically uses additional learnable components for temporal-spatial modeling, such as self-attention layers,  additional prompts, and a dedicated video decoder. Among the mentioned components, additional prompts~\cite{zhou2022conditional,khattak2023maple} are widely adapted due to their efficiency when transferring to downstream tasks without re-training the model's parameters. 


\section{Method}
\subsection{Model Architecture}
\begin{figure*}[t]
  \centering
  \includegraphics[width=\linewidth]{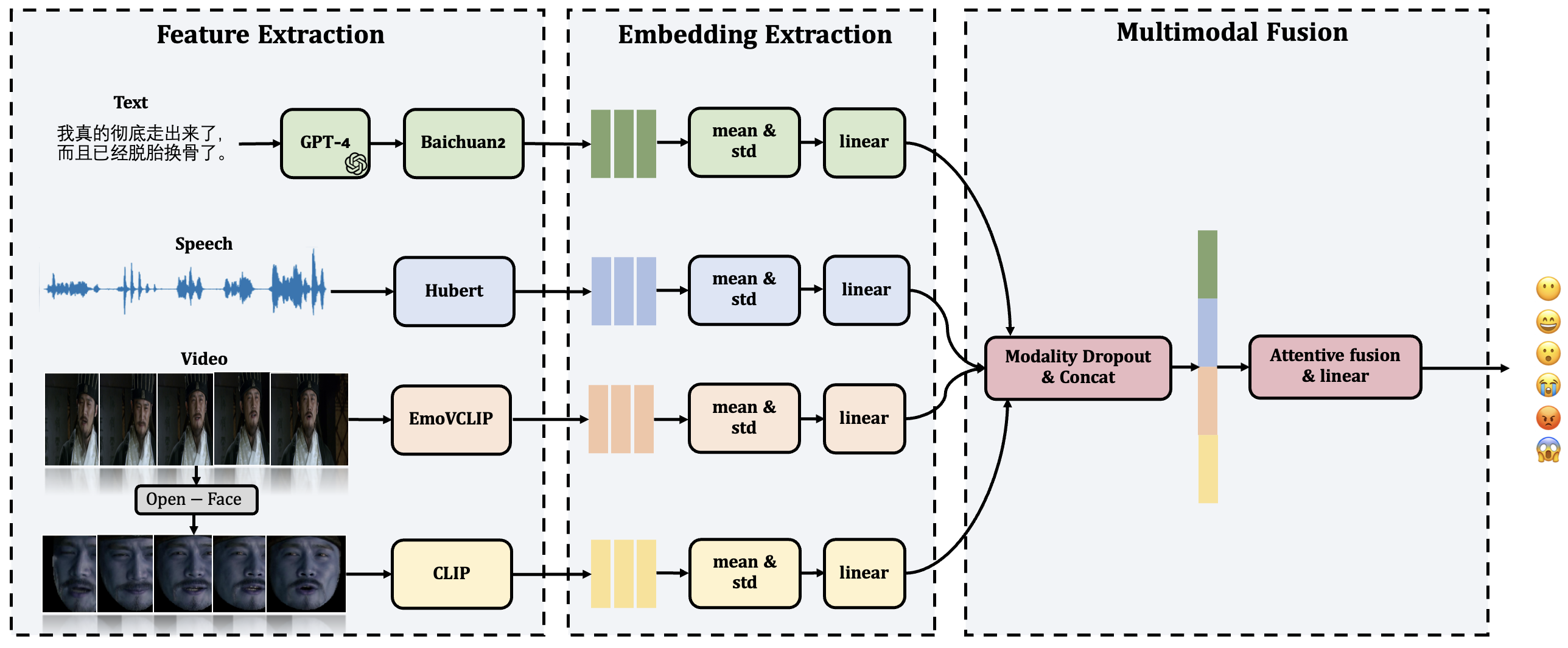}
  \caption{Framework of our proposed method. 
  }
  \label{fig1}
\end{figure*}

The overall architecture of the proposed model is illustrated in Figure.~\ref{fig1}, which consists of multiple single-modal feature extractors and a multimodal feature fusion network. In brief, the input $x \in {\{x_S,x_I,x_T,x_V\}}$ are fed into the corresponding single-modal feature extractor to obtain high-dimensional feature $f \in {\{f_S,f_I, f_T,f_V\}}$. The meanings represented by the subscripts are as follows: \textbf{$ S$} for speech, \textbf{$ I$} for image, \textbf{$ T$} for text, and \textbf{$ V$} for video.
For video modality, we propose EmoVCLIP with a strong capacity for emotion extraction. 
For image modality, we use CLIP~\cite{radford2021learning} to extract image features by slicing the video into frames of images.
For speech modality, we extract speech features by feeding the raw speech signals into Hubert~\cite{hsu2021hubert}, which is widely adapted for speech-based tasks.
For text modality, we propose to use Baichuan with input text augmented by GPT4 to explicitly obtain textual features along with the corresponding emotion states. Then two dimensional inputs $ f \in\mathbb{R}^{L, C}$  are pooled into  corresponding single-modal embeddings $e \in {\{e_S,e_I,e_T,e_V\}}$ via temporal average pooling. Finally, multimodal embeddings are concentrated in channel dimensional and fed into  
the fusion networks to output the prediction of emotion labels in \textit{neutral, anger, happiness, sadness, worry, and surprise}. Next, we will provide a detailed introduction to our proposed methods.
\subsection{EmoVCLIP}
As is discussed in ~\cite{rasheed2023fine}, although simple fine-tuning of CLIP exhibits competitive performance, it is not always feasible, particularly in downstream tasks with little available data. Research~\cite{lian2024merbench} has found that fine-tuning CLIP does not improve the performance on the MER2023 dataset but instead degrades the generalization performance. We attribute this performance degradation to the lack of labeled data and the inefficiency of the fine-tuning strategy. Inspired by Vision-Language prompt learning~\cite{rasheed2023fine}, we use prompt learning to effectively improve the generalization ability of the model on specific emotion datasets instead of fine-tuning CLIP.

As shown in Figure 2, specifically, we keep the structure and parameters of CLIP frozen, but add N prefix learnable prompt tokens at each layer of the image encoder and text encoder. For each frame of the video, the image is input into CLIP, and finally, all frames are pooled in the time dimension to calculate similarity with the text embedding
\begin{figure}[t] 
    \centering 
    \includegraphics[width=\linewidth]{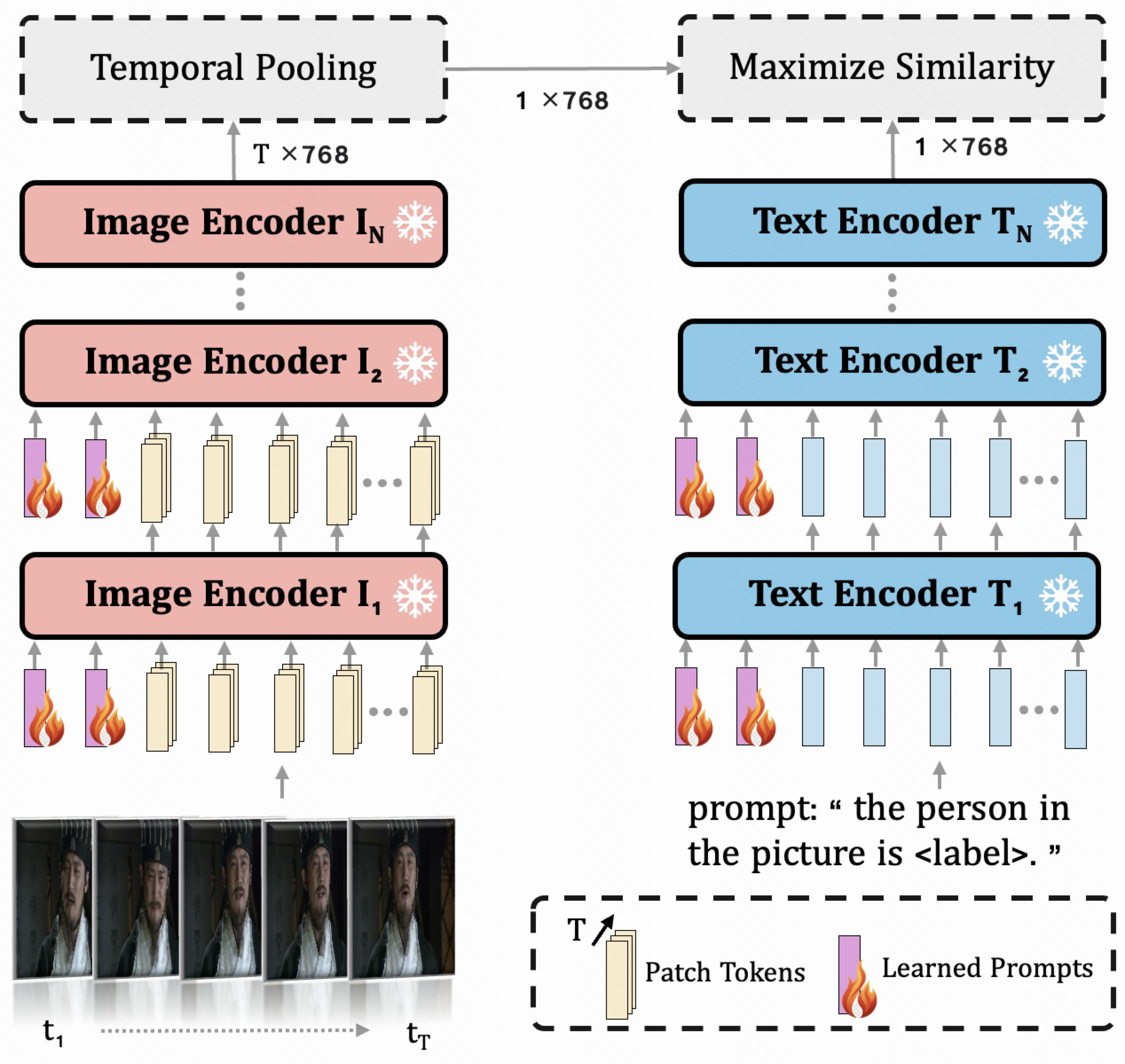} 
    \caption{Vision-Language prompt learning for EmoVCLIP. <label> in text prompts denotes ground truth emotion labels in \{neutral, anger, happiness, sadness, worry, and surprise\}.} 
    \label{fig2} 
\end{figure}

\subsection{Modality Dropout}
In u-HuBERT~\cite{hsu2022u}, modality dropout is proposed to learn modality-agnostic representations by randomly dropping certain modalities during training, thereby improving the model's generalization capabilities across various modalities. 

In this paper, we propose random modality dropout to promote the integration of features from different modalities and alleviate modality competition.
Specifically, in our model, each modality has its own extracted embeddings before multimodal fusion:  $ e_S$ for speech, $ e_I$ for image, $ e_T$ for text, and $ e_V$ for video. Embeddings are concentrated and then fed into fusion networks $g$ to predict the emotion labels. This process can be dropped as Eq.~\ref{equ:fuse}.
\begin{equation}
\label{equ:fuse}
pred=g({\mathrm{concat}(e_S,e_I, e_T,e_V))}
\end{equation}
In our method, each modality's embedding is replaced by zero vectors with a certain probability before concentration.
The real multimodal fusion can be described as Eq.~\ref {equ:fuse1}.


\begin{equation}
\label{equ:fuse1}
\left\{
\begin{array}{l} pred=g({\mathrm{concat}(e_S,e_I,e_T,e_V))}
\\e_i=\textbf{0}, i \in\{S,I,T,V\}, p=p_1
\end{array}\right.
\end{equation}
where $p_1$ denotes the modality dropout rates.

\subsection{GPT4-Baichuan}
Although Baichuan excels in Chinese-based tasks, we found that GPT4\cite{achiam2023gpt} surpasses Baichuan in emotion recognition. To enhance the emotional feature extraction capabilities of Baichuan, we have integrated GPT4's emotion extraction ability with Baichuan's Chinese language processing ability via text augment. Considering that it is impossible to obtain the feature embeddings by GPT4, we input the text and prompt words into GPT4, let GPT4 pay attention to the emotional information in the text, and sort by the possibility of six emotion labels according to the text. Finally, the output of GPT4 is spliced with the text and input into Baichuan to obtain the text information and the corresponding emotional state more clearly.

\subsection{Self-training}
Following~\cite{chen2023semi}, we adopt self-training as it exhibits promising capabilities in semi-supervised learning by effectively leveraging unlabeled data. Initially, we train a foundational multimodal emotion recognition model using labeled data. Then, we engage in an iterative self-training strategy which involves expanding the labeled dataset with pseudo-labeled data, ending with the retraining of the model. 

\section{EXPERIMENTS AND RESULTS}
\subsection{Datasets}
MER2024-SEMI provides two sets, Train \& Val with labels and Test without labels, containing 5030 and 115595 videos respectively. The final test set contains 1169 videos, including 115595 unlabeled videos. Since Train \& val is not divided into train set and validation set, similar to the baseline system, we choose to perform five-fold cross-validation, and then weigh the best results on the five validation sets equally to obtain the final result. The evaluation indicator is the weighted average F-score(WAF) defined by the challenge organizers.
\subsection{Implementation Details}
We used four feature extractors, Chinese-Hubert-large\footnote{https://huggingface.co/TencentGameMate/chinese-hubert-large}, CLIP \footnote{https://github.com/openai/CLIP}
, EmoVCLIP, and GPT4-Baichuan. The 24k speech signals extracted from the video are fed into the Chinese-Hubert-large and the output of the last four layers are added as the frame-level features of the speech modality. We used the open-face toolkit\footnote{https://github.com/TadasBaltrusaitis/OpenFace} to extract the facial features of the main faces in each frame of the video and aligned them to 128*128 pixels, and finally obtained the frame-level features. We directly input each frame of the video into EmoVCLIP without using open-face to obtain video-level features. When fine-tuning EmoVCLIP, we set the number of learned prompts of each layer to 4 and the layers are the first 12 layers of CLIP. After inputting the text and prompt into GPT4, we obtained the probability ranking of the six categories of emotion labels and then concatenated the ranking and text into Baichuan\footnote{https://huggingface.co/baichuan-inc/Baichuan-13B-Base} to obtain frame-level features. For the frame-level features, we calculated the mean and variance and concatenated them during training to obtain sentence-level features. Finally, we combined modality dropout with the attention used by the baseline to fuse different modalities.
In the self-training process, we select top k = 10 pseudo-labeled samples from each class and set the maximum number of steps Nr = 10. In each fold of 5 folds, we set the maximum number of epochs to 30 and the batch size to 64. The Adam [12] optimizer is utilized with a learning rate of 3e-4. Dropout [24] and modality dropout rate are set to 0.3.

\subsection{Results and Analysis}
Table~\ref{tab1} shows our experiments in different modalities. First, in the single modal, we try to reproduce the baseline system under the same conditions. Then we show the model results obtained by all the methods we proposed. The experiment shows that our method exceeds the best baseline result by 3\%. At the same time, to verify the effectiveness of our proposed method, we conducted an ablation experiment and compared the performance of EmoVClIP and CLIP single modal on the test set. It can be found that EmoVClIP has better emotional features than CLIP, and EmoVCLIP and CLIP have certain complementary characteristics, which means that although EmoVCLIP pays more attention to the time change information of the frame sequence after fine-tuning CLIP, it will lose some of its focus on facial features due to the lack of open-face processing. We will also show the comparison between GPT4-Baichuan and Baichuan, but there is no complementarity between GPT4-Baichuan features and Baichuan. As can be seen from the table, using modality dropout is more conducive to the fusion of different modalities than not using it, and alleviates the dependence and competition effects between modalities.

\begin{table}[t]
  \begin{center}
    \caption{Recognition results of methods with single modality}
    \label{tab1}
    \begin{tabular}{c | c|c|c} 
    \hline
    \textbf{Model} & \textbf{Feature} & \textbf{Train \& Val} & \textbf{Test} \\ \hline
    \multicolumn{4}{c}{\textbf{Audio}} \\ \hline
    Baseline & Hubert & 73.02±0.13 & 83.42±0.37 \\ 
    Ours & Hubert & 71.46 & 82.13 \\ \hline
    \multicolumn{4}{c}{\textbf{Text}} \\ \hline
    Baseline & Baichuan & 54.86±0.12 & 56.63±0.18 \\ 
    Ours & Baichuan & 53.81 & 55.44 \\ 
    Ours & GPT4-Baichuan & 56.65 & 56.84 \\ \hline
    \multicolumn{4}{c}{\textbf{Vision}} \\ \hline
    Baseline & CLIP & 66.66±0.12 & 63.27±0.29 \\ 
    Ours & CLIP & 61.96 & 60.02 \\ 
    Ours & EmoVCLIP & 68.43 & 61.08 \\ 
    Ours & CLIP+EmoVCLIP & 66.83 & 64.18 \\ \hline
    \end{tabular}
  \end{center}
\end{table}

\begin{table}[t]
  \begin{center}
    \caption{Recognition results with Multimodal. The baseline system adopts Hubert, Baichuan and CLIP as feature extractors while ours uses Hubert, GPT4-Baichuan, CLIP, and EmoVCLIP. The following models use self-training except for the baseline}
    \label{tab2}
    \begin{tabular}{c| c|c|c}
    \hline
   \textbf{Model}  & \textbf{Dropout} & \textbf{Train\&Val} & \textbf{Test}  \\ \hline
    Baseline  & 0 & 80.34±0.10 & 86.32±0.22  \\ \hline
    Ours w/o& \multirow{2}{*}{0} & \multirow{2}{*}{82.64} & \multirow{2}{*}{88.68}  \\ 
 EmoVCLIP &&&\\\hline
    Ours  & 0 & 84.25 & 89.52  \\ \hline
    Ours & 0.15 & 84.04 & 89.87  \\ \hline
    Ours  & 0.3 & 83.76 & 90.15  \\ \hline
    Ours  & 0.5 & 82.34 & 89.08 \\ \hline
    \end{tabular}
  \end{center}
\end{table}

\section{Conclusions}
We proposed a method to use prompt learning to fine-tune CLIP to adapt to emotion recognition in the video domain. We used GPT4 to strengthen Baichuan's attention to the order of sentiment classification of text. When fusing different modalities, we proposed using modality dropout to alleviate modal dependence and modal competition and used self-training to utilize unlabeled data. Finally, our model achieved a weighted F1 score of 90.15\% in the emotion recognition results of the MER2024-SEMI challenge, and we verified the effectiveness of the proposed method through ablation experiments.


\newpage
\bibliographystyle{ACM-Reference-Format}
\balance
\bibliography{sample-base}

\end{document}